\documentclass[runningheads]{llncs}
\usepackage{graphicx}

\usepackage{times}
\usepackage{latexsym}
\usepackage[numbers]{natbib}
\usepackage{tikz,lipsum,lmodern}
\usepackage[most]{tcolorbox}
\renewcommand{\figurename}{Figure}
\usepackage[T1]{fontenc}
\usepackage{graphicx}
\usepackage{float}
\usepackage{color, colortbl}
\usepackage{xcolor}
\definecolor{Gray}{gray}{0.9}

\usepackage{makecell}
\usepackage{lipsum}
\usepackage{float}
\usepackage{verbatim} %In the preamble 
\usepackage{placeins}
\usepackage{hyperref}
\usepackage{color}

\usepackage{epstopdf}
\usepackage{algorithm}
\usepackage{amsmath}
\usepackage{amssymb}
\usepackage{algorithmic}
\usepackage{algorithm}
\usepackage{subcaption}
\usepackage{pifont}
\usepackage{multirow}
\usepackage{subcaption}

\usepackage[draft,textsize=footnotesize,textwidth=15mm]{todonotes}

\usepackage[utf8]{inputenc}

\usepackage{microtype}

\usepackage{inconsolata}

\begin{document}
\title{\textit{MedSumm}: A Multimodal Approach to Summarizing Code-Mixed Hindi-English Clinical Queries}
\titlerunning{}

\titlerunning{MedSumm}
\author{Akash Ghosh\inst{1} \and
Arkadeep Acharya\inst{1} \and
Prince Jha\inst{1}\and
Aniket Gaudgaul \inst{1}\and
Rajdeep Majumdar \inst{1}\and
Sriparna Saha\inst{1}\and
Aman Chadha\inst{2,3}\thanks{Work does not relate to position at Amazon.} \and
Raghav Jain \inst{1}\and
Setu Sinha \inst{4}\and
Shivani Agarwal\inst{4}
}
% %
\authorrunning{A. Ghosh et al.}
\institute{Department of Computer Science And Engineering,Indian Institute of Technology Patna \and
Stanford University \and
Amazon GenAI\and
Indira Gandhi Institute of Medical Sciences}

\maketitle              % typeset the header of the contribution
\begin{abstract}
In the healthcare domain, summarizing medical questions posed by patients is critical for improving doctor-patient interactions and medical decision-making. Although medical data has grown in complexity and quantity, the current body of research in this domain has primarily concentrated on text-based methods, overlooking the integration of visual cues. Also prior works in the area of medical question summarisation have been limited to the English language. This work introduces the task of multimodal medical question summarization for codemixed input in a low-resource setting. To address this gap, we introduce the Multimodal Medical Codemixed Question Summarization (\textit{MMCQS}) dataset, which combines Hindi-English codemixed medical queries with visual aids. This integration enriches the representation of a patient's medical condition, providing a more comprehensive perspective. We also propose a framework named \textit{MedSumm} that leverages the power of LLMs and VLMs for this task. By utilizing our \textit{MMCQS} dataset, we demonstrate the value of integrating visual information from images to improve the creation of medically detailed summaries. This multimodal strategy not only improves healthcare decision-making but also promotes a deeper comprehension of patient queries, paving the way for future exploration in personalized and responsive medical care. Our dataset, code, and pre-trained models will be made publicly available.
% A sample of our code, dataset and pre-trained model can be found here: \url{https://anonymous.4open.science/r/MedSumm-FE55/}
\url{https://github.com/ArkadeepAcharya/MedSumm-ECIR2024}

\end{abstract}

\section{Introduction}

\textbf{Disclaimer}: The paper includes explicit medical imagery, necessary for an in-depth understanding of the subject matter.\par
Recent surveys conducted by World Health Organsisation (WHO) reveals a drastic uneven doctor to population ratio, estimating a deficit of 12 million healthcare workers by 2030. This augmented with advancements in information and communication technologies (ICTs) has experienced surge in telehealth \cite{nittari2020telemedicine}. The COVID-19 pandemic has further accelerated the utilization of the internet for healthcare services, marking a remarkable surge in the past two decades and establishing a new norm in healthcare. In this context, one of the biggest challenges that doctors face is to quickly comprehend and understand the patient's query. To solve this problem, a medical question summarizer has emerged as a vital tool to distill information from consumer health questions, ensuring the provision of accurate and timely responses. Despite previous research efforts, the unexplored opportunity of combining textual information with visual data, such as images, has been largely ignored. Visual aids play a crucial role in medical question summarization (MQS) for several reasons. A considerable segment of the population isn't well-versed in medical terminology, making it difficult to precisely convey symptoms. Additionally, certain symptoms are inherently challenging to express using only text. Patients can sometimes get mixed up when it comes to similar symptoms, like trying to tell the difference between skin dryness and a skin rash. Using a combination of text and images in medical question summaries can improve accuracy and efficiency, giving a holistic picture of the patient's current medical status. This strategy acknowledges the intricate nature of patient inquiries, where visuals like symptom photos or medical reports can offer vital insights. By prioritizing image integration, researchers and healthcare professionals can address the changing demands of contemporary healthcare communication. By improving doctor-patient communication and understanding through multimodal summarization of health questions, this work has the potential to increase access to quality healthcare and promote health equity which is one of the SDG proposed by UNESCO \footnote{\url{https://en.unesco.org/sustainabledevelopmentgoals}}.\par
Large Language Models (LLMs) \cite{kojima2023large} and Vision Language Models (VLMs) \cite{zhang2023vision} have showcased impressive capabilities in generating human-like text and multimedia content. This capability has led to their application in the medical field, primarily for specialized tasks such as summarizing chest X-rays \cite{thawkar2023xraygpt} and generating COVID-19 CT reports \cite{liu2021medical}. However, their use in summarizing medical questions that involve both text and images is a relatively unexplored area. Utilizing the zero-shot and few-shot learning abilities of these models \cite{dong2022survey} can be advantageous, particularly for tasks like multimodal medical question summarization, which often suffer from a lack of sufficient data. 

Nonetheless, there are certain limitations to consider when employing LLMs and VLMs in this context. Generic LLMs and VLMs may not possess specialized knowledge in medical domains, potentially leading to summaries that lack important details such as symptoms, diagnostic tests, and medical intricacies. On the visual side, although VLMs have excelled in typical visual-linguistic tasks, medical imaging presents unique challenges. Models like SkinGPT4 \cite{zhou2023skingpt}, which fine-tune on skin disease images and clinical notes based on MiniGPT4 \cite{zhu2023minigpt}, remain highly domain-specific. Medical images are inherently intricate, requiring a deep understanding of medical terminology and visual conventions, often necessitating input from expert medical professionals for accurate interpretation. This complexity, coupled with potential gaps in contextual comprehension, can result in models generating summaries that may be misleading or irrelevant.\par

\begin{figure}
  \centering
  \includegraphics[width=1\textwidth]{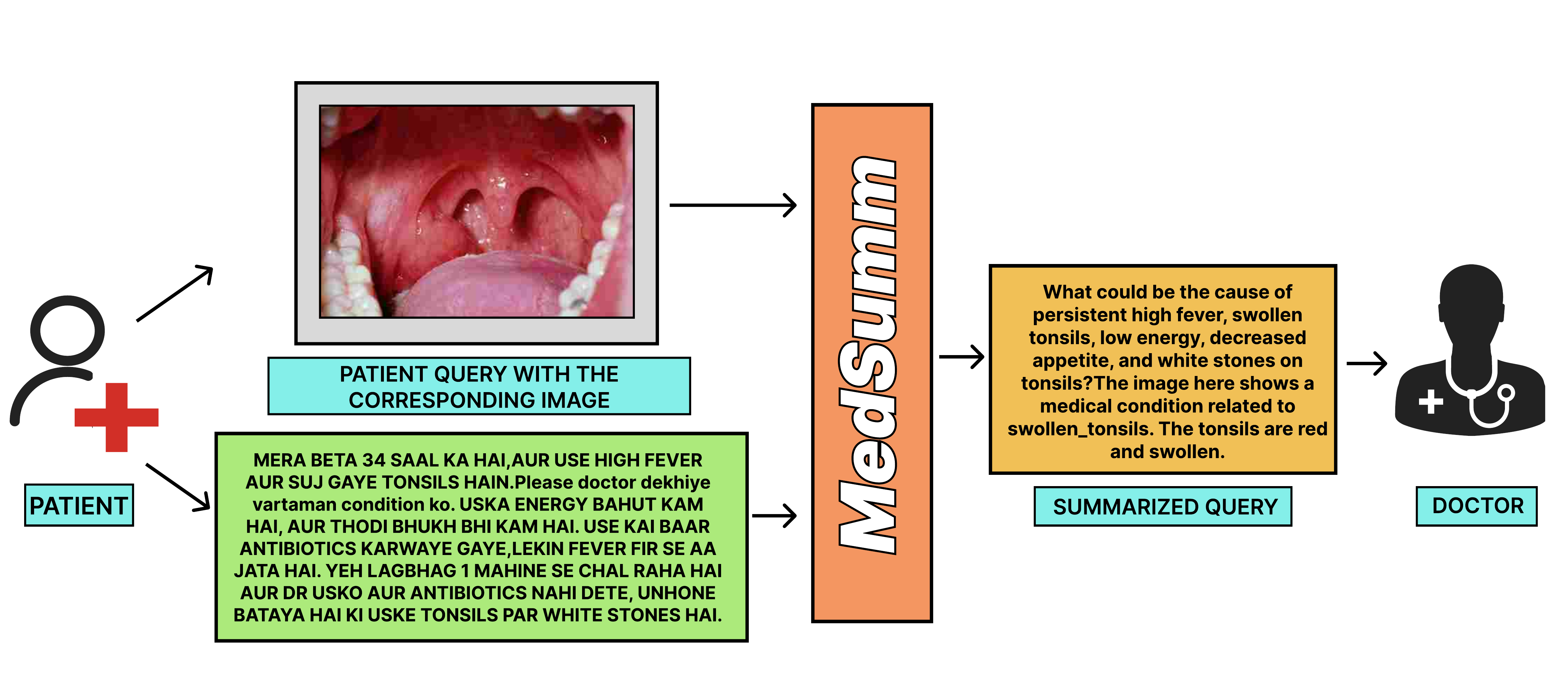}
  \caption{Pipeline demonstrating the application of MedSumm for the summarization of multimodal patient queries in a code-mixed environment, aimed at facilitating expedited comprehension of patient queries for doctors and healthcare professionals.}
  \label{fig:use_case}
\end{figure}

Also in the 21st century where more than half of the population of the globe is multilingual, people generally switch between languages in conversation\footnote{\scriptsize{\url{https://www.britishcouncil.org/voices-magazine/few-myths-about-speakers-multiple-languages}}}. Though nowadays a lot of work is going on in this exciting area of codemixing, in the clinical domain there is a dearth of good datasets. This motivated us to study our proposed task in a codemixed context.We present a curated dataset tailored to this endeavor, \textbf{\textit{MMCQS}} containing Hindi-English codemixed medical queries with their corresponding English summaries.This is the first dataset of its kind to validate this task. The dataset encompasses 3,015 medical questions along with their corresponding visual cues.
Our approach is embedded within our novel architecture known as \textit{\textbf{\textit{MedSumm}}}, a vision-language framework for medical question summarization task. This comprehensive framework operates using two primary inputs: the codemixed patient question and its associated visual cue. The methodology comprises four distinct components. They are employing pre-trained large language models to generate textual embeddings, utilizing vision encoders like ViT \cite{dosovitskiy2020image} to encode visual information, harnessing QLoRA \cite{dettmers2023qlora}, a low-rank adapter technique for efficient fine-tuning and employing a precise inference process to generate the corresponding symptom name and the synopsis. This careful, step-by-step structuring of \textbf{\textit{MedSumm}} allows it to bridge the divide between general-purpose models and the niche requirements of medical question summarization, effectively integrating textual and visual data to create precise and context-aware medical summaries.A pipeline showing the application of MedSumm has be shown in figure \ref{fig:use_case}. Our contributions can be summarized as follows:
\vspace{-0.20cm}
\begin{itemize}
    \item A novel task of \textbf{Multimodal Medical Question Summarization} for generating medically nuanced summaries.
    \item A novel dataset, \textbf{\textit{MMCQS}} Dataset, to further research in this area.
    \item A novel framework, \textbf{{\textit{MedSumm}}} that employs a combination of pre-trained language models, state-of-the-art vision encoders, and efficient fine-tuning methodologies like QLoRA \cite{dettmers2023qlora} to seamlessly integrate visual and textual information for the final summary generation of multimodal clinical questions.
    \item A novel metric \textbf{\textit{MMFCM}} to quantify how well the model captures the multimodal information in the generated summary.
\end{itemize}

\vspace{-0.6cm}

\section{Related Works}
\vspace{-0.20cm}
The following work has been relevant to following two research areas namely Medical Question Summarization and  Multimodal Summarization.\par 
\textbf{Medical Question Summarization:}
In 2019, the field of Medical Question Summarization (MQS) emerged with the introduction of the MeQSum dataset, specifically designed for this purpose by Abacha et al. \cite{abacha2019summarization}. Initial MQS research utilized basic seq2seq models and pointer generator networks to generate summaries. In 2021, a competition centered around generating summaries in the medical domain was organized, as outlined in Abacha et al.'s \cite{abacha2021overview} overview. Contestants leveraged various pre-trained models, including PEGASUS \cite{zhang2020pegasus}, ProphetNet \cite{qi2020prophetnet}, and BART \cite{lewis2019bart}. Some innovative techniques, such as multi-task learning, were employed, using BART to jointly optimize question summarization and entailment tasks \cite{mrini2021joint}. Another approach involved the use of reinforcement learning with question-aware semantic rewards, derived from two subtasks: question focus recognition (QFR) and question type identification (QTR) \cite{yadav2021reinforcement}.

\textbf{Multimodal Summarization:} In order to ensure accurate diagnosis and guidance from medical professionals, it is crucial for us to communicate our medical symptoms effectively and efficiently. One way to enhance this communication is by supplementing textual descriptions with visual cues. Previous research has demonstrated the benefits of incorporating multimodal information in various medical tasks. For instance, Tiwari et al. \cite{tiwari2022dr} highlighted how multimodal information improves the performance of Disease Diagnosis Virtual Assistants. Delbrouck et al. \cite{delbrouck-etal-2021-qiai} showed that integrating images leads to better summarization of radiology reports, while Gupta et al. \cite{gupta2022dataset} illustrated the advantages of incorporating videos in medical question-answering tasks.Kumar et al.\cite{kumar2023diving} shows how multimodal information can help in summarizing news articles. The most recent work in this domain is done by Ghosh et al.\cite{ghosh2023clipsyntel} where they incorporated CLIP with LLMs to generate the final multimodal summaries. To the best of our knowledge, our work represents the first attempt to address the task of question summarization in the medical domain, particularly within a codemixed multimodal context.
\vspace{-0.25cm}
\section{MMCQS Dataset}
\vspace{-0.15cm}
\subsection{Data Collection}
\begin{figure*}
  \centering
  \includegraphics[width=1\textwidth]{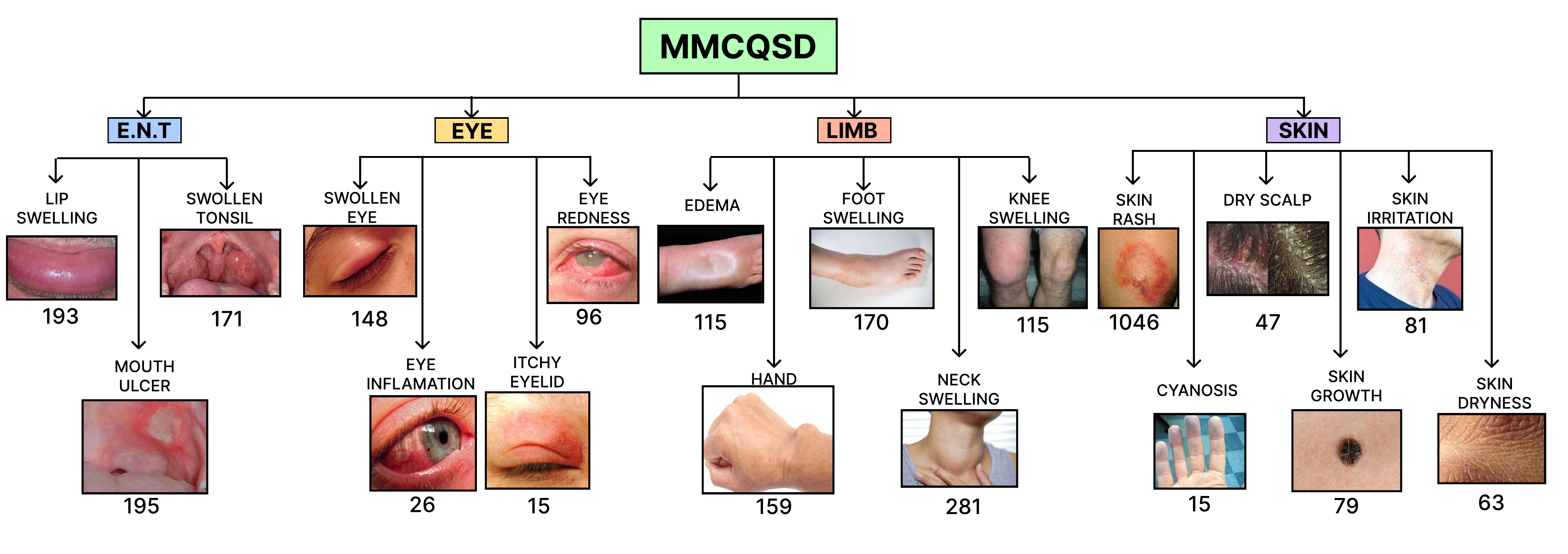}
  \caption{Broad categorization of medical disorders in the MMCQS Dataset (MMCQSD). The number of data points corresponding to each category has been provided under each category in the above figure.}
  \label{fig:ann}
\end{figure*}

\vspace{-0.10cm}
Prior to this work, there was no multimodal codemixed question summarization dataset available in the healthcare domain with textual questions and corresponding medical images. We used the HealthCareMagic Dataset, derived from MedDialog data, with 226,395 samples, after removing 523 duplicates.
We were led by medical doctors who were also the co-authors of the paper,  identified 18 medical symptoms that are hard to convey through text, and divided into four groups: ENT, EYE, LIMB, and SKIN. The entire categorization is shown in Figure -\ref{fig:ann}. We selected symptoms and obtained images using the Bing Image Search API\footnote{\url{https://www.microsoft.com/en-us/bing/apis/bing-image-search-api}}, which were verified by medical students. Our dataset features instances mentioning body parts in questions and summaries. We used FlashText\footnote{\url{https://pypi.org/project/flashtext/1.0/}} for term matching and Textblob4\footnote{\url{https://textblob.readthedocs.io/en/dev/}} to correct misspellings, resulting in a final dataset of 3,015 samples for multimodal summarization.

\subsection{Data Annotation}
\begin{comment}
We randomly selected 100 samples from the dataset, each containing a patient's inquiry and its corresponding summary. These samples were provided to the medical experts, who are coauthors of this paper, to develop annotation guidelines and create the final annotated samples for annotation training. After a few rounds of brainstorming it is decided that the annotation process will be divided into three broad categories. These include (A) Incorporation of Visual Cues (B) Updation of Golden summaries, and (C) Hindi-English codemixed Conversion.Each of these categories is elaborated below.\par
\end{comment}
We randomly selected 100 samples from the dataset and provided them to the medical experts, who are co-authors of this paper. They developed annotation guidelines for the annotation training process, divided into three categories: (A) Incorporation of Visual Cues, (B) Updating Golden Summaries, and (C) Hindi-English Codemixed Conversion.\par
\vspace{0.15cm}
\textbf{A. Incorporation of Visual Cues:} 
\begin{comment}
The methodology through which the medical expert introduced annotated these samples is explained with the given example. For example, suppose the patient is complaining regarding something that has happened near their tonsils, but they are not able to name the exact medical disorder they are suffering from. But their textual reference gave enough context to understand that the patient is suffering from the disorder named \textbf{swollen-tonsils}. After understanding this, the medical expert adds statements like \textbf{\textit{Please see what happened to my tonsils in the image below}} in the context and accordingly adds a suitable visual image representation of the disorder. In this way, visual medical signs are incorporated into the textual medical question.
\end{comment}
The methodology involved medical experts annotating samples with visual cues. For instance, when a patient's description hints at a medical issue near their tonsils, the expert adds relevant statements and visual images to clarify the condition, such as adding \textbf{\textit{Please see what happened to my tonsils in the image below}}. This process integrates visual medical signs into the textual medical question

\par
\vspace{0.15cm}
\textbf{B. Updating the Golden Summaries:}
\begin{comment}

The medical expert deduced that the conventional golden summaries did not align well with the multimodal queries, prompting the need for corresponding updates. To address these issues, the medical experts themselves modified these 100 samples by revising the questions, incorporating the visual information, and adapting the golden summaries accordingly. Few samples of the updated golden summaries are  shown in the qualitative analysis section.
\end{comment}

The medical experts recognized a misalignment between conventional golden summaries and multimodal queries, leading them to update the summaries. They revised the questions, incorporated visual information, and adapted the golden summaries for the 100 samples. These 100 samples are used for annotation guidelines for the annotators. 
% \acil(Some updated summaries are presented in the qualitative analysis section.)

\par
\begin{figure}
  \centering
  \includegraphics[width=0.8\textwidth]{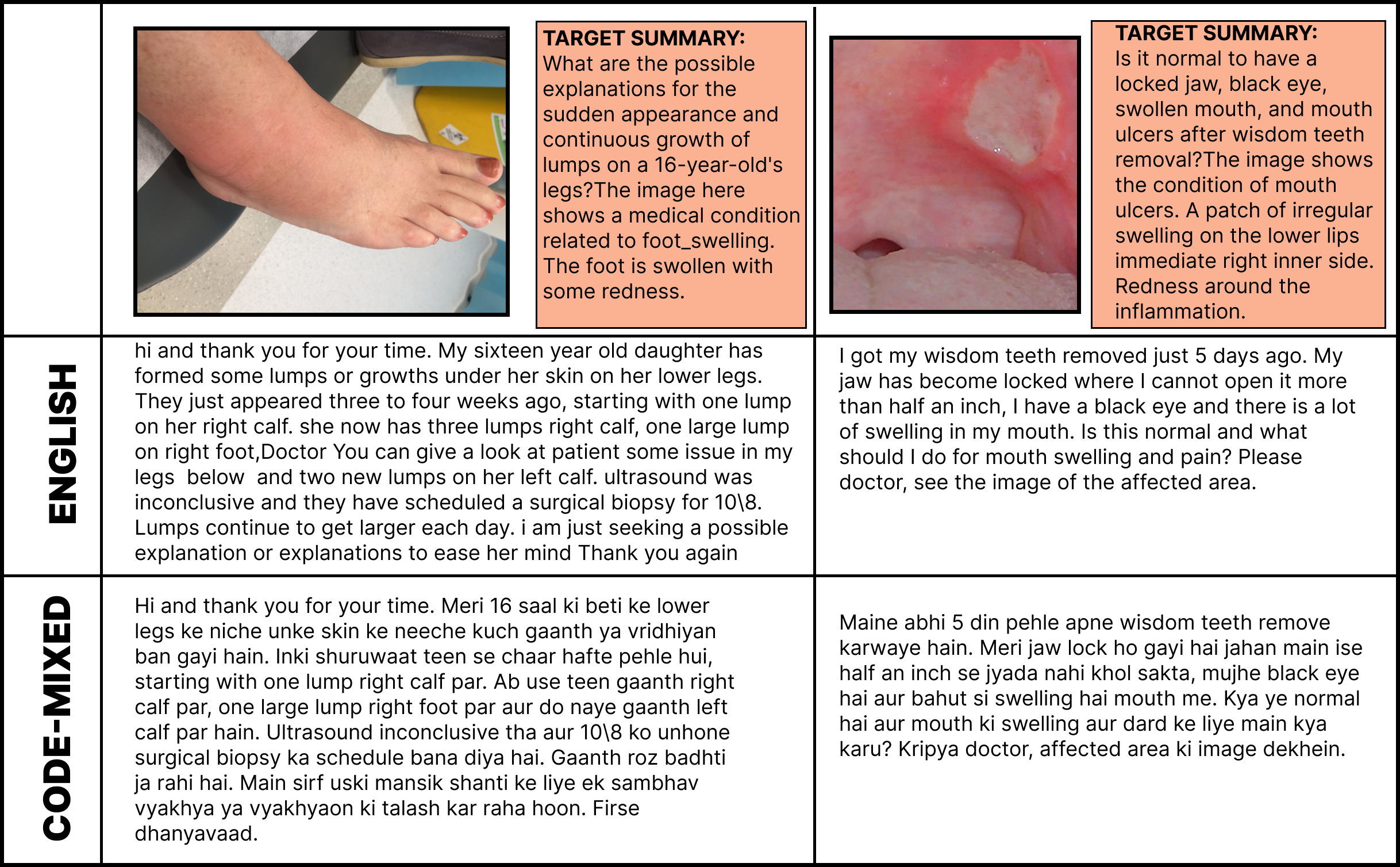}
  \caption{Sample instance of the MMCQS Dataset with the corresponding English query and updated golden summary (target summary).}
  \label{fig:data_samples}
\end{figure}

\textbf{C. Hindi-English Codemixed Annotation:}
\begin{comment}
For this task, the medical experts annotated around 50 samples. Recent work shows that GPT-3.5 turbo performed really well in codemixed generation task \cite{yong2023prompting}. So we experimented with GPT-3.5 in a few shot prompting techniques with the annotated samples \cite{yong2023prompting} and used the below prompt to generate the Hinglish (Hindi + English) mixed version of the English text. The final Hinglish text has a code-mixing index 
\cite{das2014identifying} of 30.5. This indicates that the data has a decent mix of Hindi and English words in the generated samples. The prompt used to generate the final Hinglish translation  is presented below:\par
\end{comment}
For this task, around 100 samples were annotated by medical experts. We conducted experiments using GPT-3.5 with few-shot prompting techniques on annotated samples due to its strong performance in generating codemixed text \cite{yong2023prompting}. The goal was to generate a Hinglish (Hindi + English mixed) version of the English text. The generated Hinglish text had a code-mixing index \cite{das2014identifying} of 30.5, indicating a good mix of Hindi and English words. The specific prompt used is presented below. 80 samples were tested, with annotators rating code-mixing out of 5. After verification with the golden test examples, we get an average score of 3.2 out of 5 indicating the codemixed data is of reasonable quality. An instance of the Hindi-English Codemixed annotated data along with their corresponding English query and updated golder summaries has been shown in figure \ref{fig:data_samples} 
\begin{tcolorbox}[colback=blue!5!white,colframe=blue!75!black,title=Prompt used for codemixed text generation]

  \textbf{\textit{You are a linguistic expert whose task is to convert the English passages into corresponding Hinglish codemixed ones.\textlangle Labelled Examples\textrangle: English: \{text\}  Hinglish: \{text\} .
 \hspace{30 mm}  Given the English passage: \{text\}, convert it into the corresponding Hinglish passage shown in the  \textlangle Labelled Examples\textrangle.}}

\end{tcolorbox}

\vspace{-0.5cm}

\subsection{Annotation Training and Validation}
\vspace{-1mm}
To ensure high-quality annotation aligned with ethical guidelines, we engaged three postgraduate medical students proficient in Hindi and English. They received guidance and conducted sessions with medical experts to clarify task-related doubts. Breaks were provided every 45 minutes during the approximately 4-month, 90-training session annotation process\footnote{The medical students were compensated through gift vouchers and honorarium amount in lines with \url{https://www.minimum-wage.org/international/india}.}.

In the validation phase, the dataset was divided into three parts, and annotators assessed fluency, adequacy, informativeness, and persuasiveness, maintaining inter-annotator agreement using the Cohen-Kappa coefficient. Annotators reviewed and discussed inaccuracies in each other's sets, leading to improved quality scores from the initial (fluency = 3.5, adequacy = 3.01, informativeness = 2.85, persuasiveness = 2.25) to the final phase (fluency = 4.8, adequacy = 4.7, informativeness = 4.1, persuasiveness = 4.45). The annotators achieved a kappa coefficient of 0.75, indicating annotation consistency.
\vspace{-0.5cm}
\vspace{-1cm}
\vspace{1.0cm}
\section{Methodology}
\vspace{-0.20cm}
\subsection{Problem Formulation}
\begin{figure}
  \centering
  \includegraphics[width=1\textwidth]{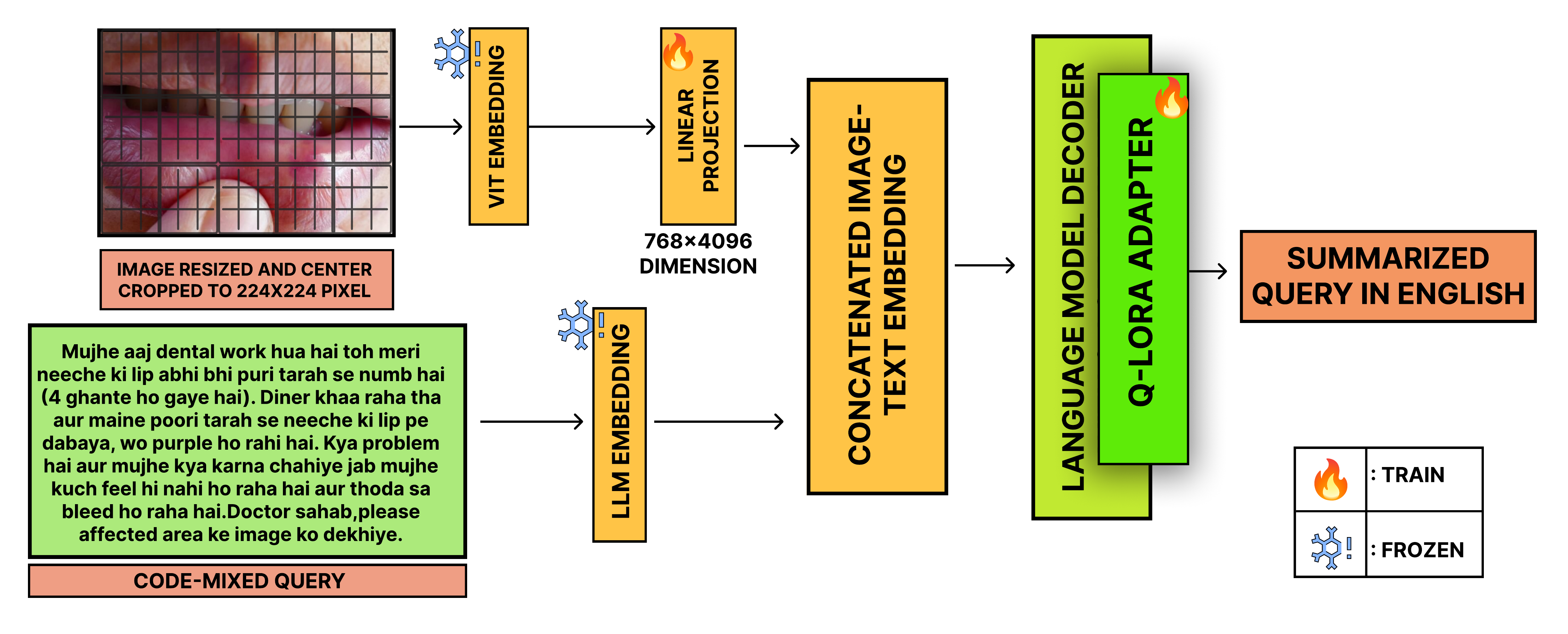}
  \caption{Model structure of MedSumm showing the different stages of our proposed architecture. The frozen and trainable layers have also been indicated.}
  \label{fig:pipeline}
\end{figure}

Each data point comprises a patient's textual inquiry, denoted as $Q$, along with an accompanying image, represented as $I$, illustrating the medical issue or concern the patient is seeking assistance within their text. The ultimate objective is to generate a natural language sequence $Y$ represented as:\begin{equation}
Y = \{S,T\}
\end{equation}
where  $S$ represents the succinct summary that incorporates insights from both the textual and visual modalities and  $T$ represents a brief note about the medical condition associated with the visual cue. Figure \ref{fig:pipeline} shows a diagrammatic representation of our proposed architecture MedSumm. MedSumm has 3 main stages namely: (i) \textbf{Question and Visual Symptom Representation}, (ii) \textbf{Adaptation Methods}, and (iii) \textbf{Inference }.

\subsection{Textual and Visual Representation}
\vspace{-0.15cm}
We've utilized two distinct categories of data to represent a patient-clinician interaction: the patient's question in textual form and visual elements that the patient shared alongside the query.\par
\vspace{0.15cm}
\textbf{Question Representation: }
In this context, each patient's query is presented as a text passage in which they explain their medical concerns to the doctor to receive relevant feedback. Recent advancements in language models, particularly decoder-based ones like LLaMA \cite{touvron2023llama1}, GPT-3\cite{brown2020language}, have demonstrated superior performance in encoding textual presentations compared to encoder-based models like BERT\cite{devlin2018bert}.
This technique involves employing Large  Language Models (LLMs) for next-word prediction, from which we generate sentence embeddings by extracting the hidden vectors corresponding to the final word. Various LLMs, such as Vicuna \cite{zheng2023judging} Llama2 \cite{touvron2023llama2}, FLAN-T5 \cite{chung2022scaling}, Mistral-7B \cite{jiang2023mistral} and Zephyr-7B \cite{tunstall2023zephyr} 
, are utilized for this purpose. During this process, the text is tokenized into smaller units, and each token is ultimately transformed into a 4096-dimensional embedding. It is noteworthy that pre-trained models of these LLMs are employed in this task.

\textbf{Visual Representation: } 
We employ  Vision-transformers ViT\cite{dosovitskiy2020image}, for this task. These models take raw images as input and transform them into embeddings of size 768. Furthermore, we integrate a linear projection mechanism using a fully connected layer to map these 768-dimensional visual embeddings into a shared textual embedding space. This unified vision-language embedding is then fed into the decoder of the language model. Notably, only the linear projection layer is trainable in this process. This linear projection layer not only helps to extract meaningful information from the vision encoder but also helps in multimodal information fusion, thus producing richer information embeddings that can be effectively learned and finetuned.

\begin{comment}
    
This linear projection serves a dual purpose: firstly, it extracts valuable information from the vision encoder's embeddings, and secondly, it allows the static language model to adapt to new data by learning and fine-tuning accordingly.
\vspace{-0.8cm}
\end{comment}
\subsection{Adaptation Methods}
\vspace{-0.2cm}
One of the biggest challenges of LLM-based techniques is fine-tuning them to domain-specific tasks as they are mostly bounded by resource constraints. There are several techniques that have been developed to overcome this problem namely prompt tunning,in-context learning, and low-rank adaptation techniques like LoRA \cite{hu2021lora} and QLoRA \cite{dettmers2023qlora}. For this work, we are going with the latest and most parameter-efficient technique which is QLORA. QLoRA is a more memory-efficient version of LoRA providing 4-bit quantization to enable fine-tuning of larger LLMs using the same hardware constraints.
\begin{comment}
LORA performs gradient descent on only the injected weights while keeping the other weights in a frozen state and thus it provides a significant advantage in comparison to training the model from scratch. 
\end{comment}
\vspace{-0.5cm}
\subsection{Inference}
\vspace{-0.2cm}
The inference module is an LLM that receives the attended multi-modal fusion representation vector, which combines the patient's textual query with visual cues. This module generates the corresponding symptom name and summary using a next-token prediction approach. In this method, the language model calculates probabilities for all tokens in its vocabulary, and these probabilities are used to predict the next predicted token. We use the following LLMs namely Llama-2, Mistral-7B, Vicuna , FLAN-T5 and Zephyr-7B  for  the final summary generation
% The distribution $p_{\theta}(y)$ is represented by the following equations:

% \begin{equation}
% \log p_{\theta}(y) = \sum_{l} \log p_{\theta}(y_{l}|y_{1},y_{2},...,y_{l-1})
% \end{equation}

\section{Experiments and Results}
\vspace{-0.25cm}
To study the task of summarization of complex code-mixed medical queries we perform a meticulous fine-tuning and evaluation of both uni-modal and our multimodal model \textit{\textbf{MedSumm}} on our \textit{\textbf{MMCQS}} dataset. We have allocated 80\% of the dataset for training, 5\% for validation, and 15\% for testing our fine-tuned models. We have leveraged the most recent and popular open source LLMs like Vicuna\cite{zheng2023judging} Llama2 \cite{touvron2023llama2}, FLAN-T5 \cite{chung2022scaling}, Mistral-7B \cite{jiang2023mistral} and Zephyr-7B \cite{tunstall2023zephyr} in both unimodal and multimodal (MedLLMSumm) setup.We utilized ROUGUE \cite{lin2004rouge}, BLEU \cite{papineni2002bleu}, BERT score \cite{zhang2019bertscore}, METEOR \cite{banerjee2005meteor} as automatic evaluation metrics\footnote{To maintain uniformity in the results post-processing like removing extra spaces, repeated sentences are performed.}.  For the purpose of human evaluation, we collaborate with a medical expert and a few medical postgraduate students. We have identified four distinct and medically nuanced metrics
for this evaluation: clinical evaluation score, factual recall \cite{abacha2023investigation}, hallucination rate \cite{abacha2023investigation}, and 
our \textbf{\textit{MMFCM}} metric.\par

\textbf{Automatic Evaluation:} The automatic evaluation results in Table \ref{tab:quant} reveal that \textbf{MedSumm} consistently outperforms all other LLMs across all defined metrics. This demonstrates the significant advantage of incorporating visual cues for generating more nuanced summaries. In unimodal settings, Mistral outperforms the rest, while Zephyr, as the most recent LLM, demonstrates balanced performance in both unimodal and multimodal scenarios. In the multimodal setting, LLAMA-2 takes the lead, closely followed by Vicuna in terms of performance.FLAN-T5's performance was the weakest across all metrics. This led us to the conclusion that decoder-only architectures perform better than encoder-decoder models like FLAN T5.\par
\begin{table*}[]
\centering
\scalebox{0.8}{%
\begin{tabular}{c|c|ccc|cccc|c|c|}
\cline{2-11}
 &
  \multirow{1}{*}{\textbf{Model}} &
  \multicolumn{3}{c|}{\textbf{ROUGE}} &
  \multicolumn{4}{c|}{\textbf{BLEU}} &
   \multicolumn{1}{c|}{\textbf{BERTScore}}  &  \multicolumn{1}{c|}{\textbf{METEOR}} \\ \cline{3-9}
                                            &                & \textbf{R1}    & \textbf{R2}   & \textbf{RL}    & \textbf{B1}    & \textbf{B2}    & \textbf{B3}    & \textbf{B4}    &  &  \\ \hline
\multicolumn{1}{|c|}{\multirow{2}{*}{\rotatebox[origin=c]{90}{{\parbox{2cm}\centering \textbf{\tiny LLAMA-2}}}}}  &  Unimodal(only texual query)&  39.92  & 19.57  &  33.9  & 28.9  & 18.05 & 11.76 & 9.97 & 0.74 & 34.81 

  \\
\multicolumn{1}{|c|}{}   & & & & & & & & & &                        \\
\multicolumn{1}{|c|}{} & MedSumm  & \textbf{46.75} & \textbf{25.59} & \textbf{38.41} & \textbf{32.50} & \textbf{22.55} & \textbf{17.56} & \textbf{14.88} & \textbf{0.80} & \textbf{35.74}
                      \\  \hline \hline

\multicolumn{1}{|c|}{\multirow{2}{*}{\rotatebox[origin=c]{90}{\textbf{\small Mistral}}}} & Unimodal(only texual query) & 37.58  & 17.14  & 29.01  & 23.67  & 14.46 & 9.42 & 6.15 & 0.77 & 41.68 

  \\

\multicolumn{1}{|c|}{}   & & & & & & & & & &                          \\
\multicolumn{1}{|c|}{}  & MedSumm  &  \textbf{42.54} & \textbf{23.81} & \textbf{34.85} & \textbf{27.21} & \textbf{19.04} & \textbf{14.89} & \textbf{12.22} & \textbf{0.76} & \textbf{36.00}
  \\ \hline \hline
%\multicolumn{1}{|c|}{\multirow{4}{*}{Falcon}} &
%  OCR &
%  13.96 &
%  1.51 &
%  9.24 &
%  35.59 &
%  25.28 &
%  12.82 &
%  6.9 &
%  20.1475 &
%  12.66952956 &
%  81.27 \\
% \multicolumn{1}{|c|}{}                 %     & OCR + MiniGPT4 & 3.28  & 0.55 & 2.59  & 3.65  & 3.44  & 2.59  & 1.72  & 2.85    & 2.713786765  & 80.2  \\
%\multicolumn{1}{|c|}{}                      & OCR + VLmeme   & 7.07  & 0.7  & 5.2   & 11.62 & 9.92  & 5.86  & 3.27  & 7.6675  & 6.197163396  & 79.02 \\
%\multicolumn{1}{|c|}{}                      & MemeGuard      & 6.63  & 0.86 & 4.83  & 9.01  & 8.12  & 5.33  & 3.22  & 6.42    & 5.51264      & 79.02 \\ \hline \hline
\multicolumn{1}{|c|}{\multirow{2}{*}{\rotatebox[origin=c]{90}{\textbf{\small Vicuna}}}} &  Unimodal(only textual query)& 38.64  & 21.55  & 32.9  & 24.12  & 15.88 & 11.09  & 9.13 & 0.75 & 32.25 
\\

\multicolumn{1}{|c|}{}    & & & & & & & & & &                      \\
\multicolumn{1}{|c|}{}   & MedSumm &  \textbf{45.09} & \textbf{26.11} & \textbf{37.05} & \textbf{27.58} & \textbf{19.98} & \textbf{16.44} & \textbf{14.33} & \textbf{0.80} & \textbf{32.10}
 \\ \hline \hline
\multicolumn{1}{|c|}{\multirow{2}{*}{\rotatebox[origin=c]{90}{\textbf{\tiny FLAN-T5}}}} & Unimodal(only textual query) & 36.3  & 17.22  & 29.81  & 22.1  & 12.14  & 8.51  & 6.15  & 0.67  &  28.25 
\\
\multicolumn{1}{|c|}{}   & & & & & & & & & &                       \\
\multicolumn{1}{|c|}{}   & MedSumm & \textbf{41.5} & \textbf{23.02} & \textbf{33.96} & \textbf{26.2} & \textbf{18.04} & \textbf{14.12} & \textbf{11.8} & \textbf{0.74} & \textbf{35.09}

\\ \hline \hline
\multicolumn{1}{|c|}{\multirow{2}{*}{\rotatebox[origin=c]{90}{\textbf{\small Zephyr}}}} & Unimodal(only textual query)  & 36.67  &  17.54  & 30.12  & 22.93  & 14.01  & 8.97  & 5.80  & 0.70  &  35.32  
  \\

\multicolumn{1}{|c|}{} & & & & & & & & & &                       \\
\multicolumn{1}{|c|}{}                      &  MedSumm      & \textbf{44.55} & \textbf{25.37} & \textbf{34.97} & \textbf{27.05} & \textbf{19.48} & \textbf{15.84} & \textbf{13.63} & \textbf{0.77} & \textbf{33.37}
  \\ \hline
\end{tabular}%
   }
\vspace{0.2cm}
\caption{Performance of various \textit{MedSumm} models and corresponding unimodal baselines, evaluated using automatic metrics with different  LLMs.}
\label{tab:quant}
\end{table*}

\textbf{Human Evaluation: }
\begin{comment}
The human evaluation was done by a team of medical students led by a doctor. The team was
given 10 \% of the dataset (selected at random) for evaluation purpose and was asked to rate the summaries generated by the uni-modal models and the multi-modal models. The following metrics are used for the evaluation: \textbf{(1) Clinical Evaluation Score:} The doctor and his team were asked to rate the summaries between 1 (poor) and 5 (good) based on their overall relevance, consistency, fluency, and coherence. \textbf{(2) Multi-modal fact capturing metric (MMFCM):}
We propose a new metric to evaluate how well a model incorporates relevant medical facts and identifies the correct disorder in a multimodal setup. MMFCM is calculated by considering the facts extracted from the medical golden summary and assessing whether they are correctly incorporated in the generated summary. The metric accounts for (1) The ratio of correct facts in the summary to the total number of relevant medical facts in the target summary. (2) Additional scores based on the accuracy of the disorder's detection, with values ranging from +2 for fully correct to -1 for incorrect identification. See Algorithm 1 for the detailed algorithm. x
\end{comment}
Existing automatic evaluation metrics can provide a misleading assessment of summary quality in the medical domain, where a single incorrect detail or omission of vital information can be perilous.
A team of medical students, led by a doctor, conducted a human evaluation using a random 10\% of the dataset to rate summaries from uni-modal and multi-modal models. Evaluation metrics include: (1) Clinical Evaluation Score for relevance, consistency, fluency, and coherence (rated 1 to 5), (2) Multi-modal Fact Capturing Metric (MMFCM) assessing the incorporation of medical facts and disorder identification, and (3) Medical Fact-Based Metrics, which employ Factual Recall and Omission Recall to measure the capture of medical facts compared to gold standard summaries. See Algorithm 1 for the detailed algorithm. Table \ref{tab:human} provides a comparison of both the uni-modal models and multi-modal models, underscoring the advantage of multimodal models over unimodal models for our task. This claim is also seconded by the improvement in the Factual Recall, Hallucination Rate, and our proposed MMFCM metrices.\par
\vspace{-0.5cm}

\begin{table}[]
\centering
\scalebox{0.65}{%
\renewcommand{\arraystretch}{1.78}
\begin{tabular}{cc|cccc|}
\cline{2-6}
\multicolumn{1}{c|}{}                          & \textbf{Models}     & \textbf{Clinical-EvalScore} & \textbf{Factual Recall} & \textbf{Hallucination Rate} & \textbf{MMFCM Score} \\ \cline{2-6}
\multicolumn{1}{c|}{}    &  LLAMA-2(U)     &   3.1            &         0.34         &          0.35          &         NA            \\

\multicolumn{1}{c|}{\multirow{7}{*}} & \textit LLAMA-2(M) &  3.52           &        0.35         &           0.32           &        0.42              \\
\cline{2-6}

\multicolumn{1}{c|}{}                         &  Mistral-7B(U) &  3.41           &        0.36        &           0.29           &         NA              \\

\multicolumn{1}{c|}{}                         & Mistral-7B(M) &   3.43           &        0.36        &           0.3           &         0.41              \\

\cline{2-6}

\multicolumn{1}{c|}{}                         & Vicuna-7B(U) &  3.32            &        0.33       &           0.33           &           NA               \\

\multicolumn{1}{c|}{}                         & Vicuna-7B(M) &   3.45            &        0.34        &          0.36           &            0.41             \\

\cline{2-6}

\multicolumn{1}{c|}{}                         & FLAN-T5(U) &    2.8             &          0.28        &            0.31            &        NA             \\

\multicolumn{1}{c|}{}                         & FLAN-T5(M)  &   3.1             &          0.31       &            0.33            &        0.34            \\

\cline{2-6}

\multicolumn{1}{c|}{}                         & Zephyr 7B $\beta$(U) &   3.48            &          0.38        &           0.26            &        NA               \\

\multicolumn{1}{c|}{}                         & Zephyr 7B $\beta$(M)  &  3.54            &          0.36        &           0.29            &         0.44             \\

\hline

\multicolumn{2}{|c|} {\textbf{Annotated Summary}}               &   \textbf{4.1}     & \textbf{0.88}        & \textbf{0}               &   \textbf{0.87}              \\ \hline
\end{tabular}%
}
\vspace{0.4cm}
\caption{Human evaluation scores of various unimodal and multimodal models across different metrics.}
\label{tab:human}
\end{table}
\par

\begin{comment}
\begin{algorithm}
	\caption{MMFCM Method }
	\begin{algorithmic}
		\Require{$F_{m}=\{fact_{m,1},fact_{m,2}\dots fact_{m,n-1},fact_{m,n}$\} } \textbackslash \textbackslash Relevant Medical facts from golden summary 'm'.\newline
  
        $Sf_{m}$ =$\{{Summfact_{m,1},\dots,Summfact_{m,n}}\} $ \newline
        \textbackslash \textbackslash Relevant Medical facts of summary of query 'm'.\newline
        
  	\Ensure{$\#CorrectFacts_{m}$=  $\lvert F_{m} \cap Sf_m \rvert$ \textbackslash \textbackslash  Number of correct medical facts in each summary} \newline 
 
   \State \textbf{if} $ \left ({Correct Medical Disorder phrase \in \{ F_{m} \cap Sf_m \} } \right )$: \newline
      \State  \#CorrectFacts_{m} +=2 \newline
    \State \textbf{else if} $\left ({Partially correct disorder phrase \in \{ F_{m} \cap Sf_m \} } \right)$: \newline
      \hspace{2cm}   \State \#CorrectFacts_{m} +=1 \newline
    \State \textbf{else if }$\left ( {Incorrect disorder phrase \in \{ F_{m} \cap Sf_m \} }\right)$: \newline
      \hspace{2cm}   \State \#CorrectFacts_{m} +=-1 \newline
    \State \textbf{else}: \newline
         \State  \#CorrectFacts_{m} +=0 \newline
%    \If{$i\geq 5$} 
%     \State $i \gets i-1$
% \Else
%     \If{$i\leq 3$}
%         \State $i \gets i+2$
%     \EndIf
% \EndIf 
  
   Result = \#CorrectFacts_{m}/${\lvert F_{m} \rvert}$ \newline
      \State \textbf{if} $ \left ({Result<1} \right )$:     
       Result \newline
    \State \textbf{else}: 
         \Return 1
\end{algorithmic}
\end{algorithm}
\end{comment}
% \vspace{-0.9cm}

\vspace{0.5cm}

\begin{comment}
Summarization in healthcare necessitates strong ethical considerations, particularly regarding safety, privacy, and potential bias. To address these concerns in our project with the MMCQS dataset, we implemented several proactive measures. We collaborated closely with medical professionals and also obtained IRB approval to ensure ethical rigor and patient privacy. We rigorously followed legal and ethical guidelines\footnote{\url{https://www.wma.net/what-we-do/medical-ethics/declaration-of-helsinki/}} during dataset validation, integration of images, and annotation of summaries. Medical experts were engaged throughout the process, providing validation and correction of the dataset and also validating the outputs of the models. The proposed dataset is based on the original HealthcareMagic
Dataset,i.e., the medical questions/samples are taken from this dataset. The incorporation of multimodality and codemix annotation into the task is done under the full supervision of a medical professional. Additionally, we ensured user privacy by not disclosing identities.
\end{comment}
\begin{figure}
  \centering
  % \begin{adjustbox}
  \includegraphics[width=1\textwidth]{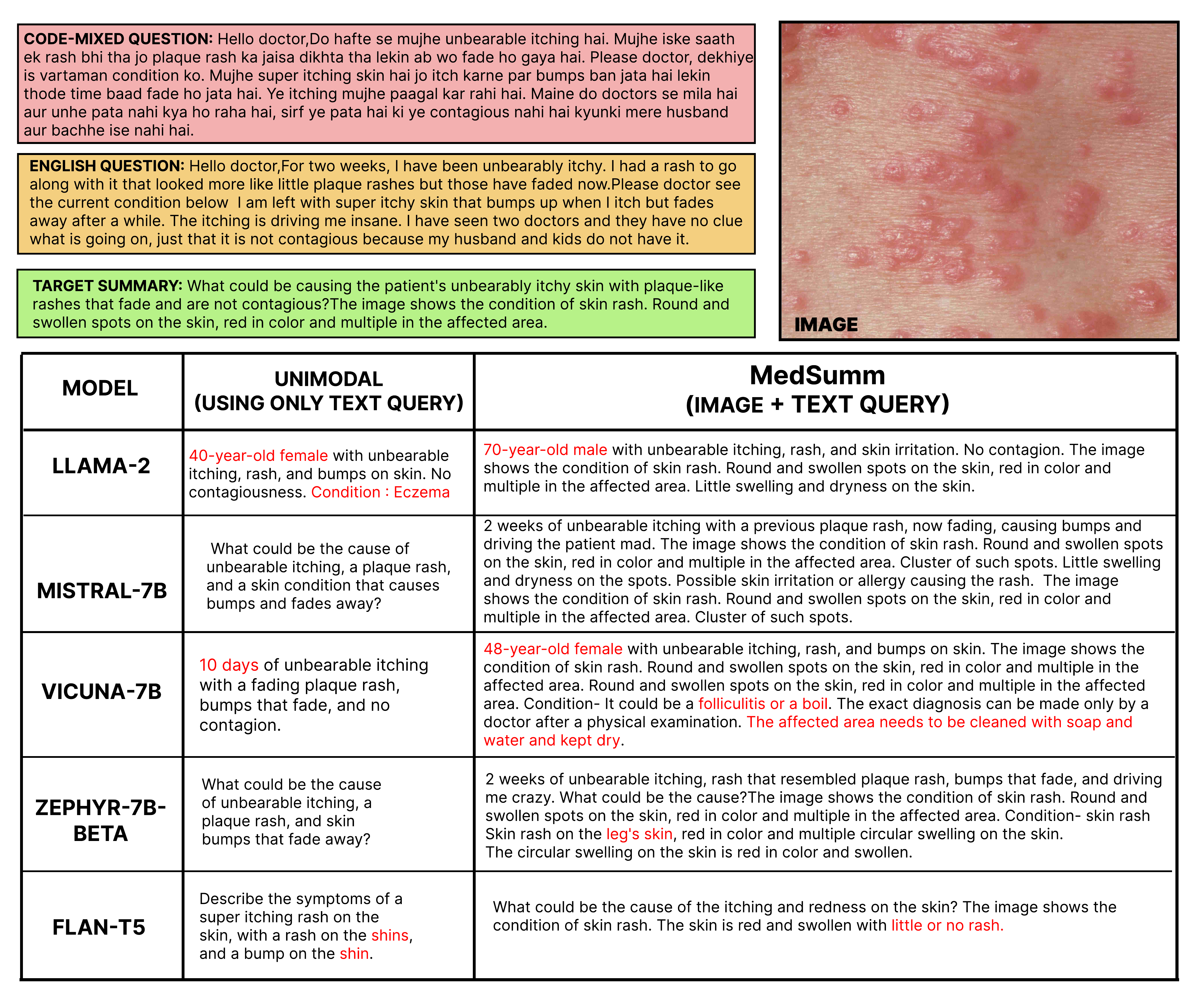}
  % \end{adjustbox}
  \caption{Samples summaries generated by the LLMs in unimodal and multimodal setting. The hallucinated phrases has been shown in red.}
  \label{fig:quali-ex1}
\end{figure}

\vspace{-0.5cm}

\begin{algorithm}
   \caption{MMFCM Method }
	\begin{algorithmic}

\REQUIRE{$F_{m}=\,\{\,\,fact_{m,1}\,,\,fact_{m,2}\,\dots \,fact_{m,n-1}\,,\,fact_{m,n}\,$\,\}\, }\newline \textbackslash \textbackslash Relevant Medical facts from golden summary of query 'm'.

$Sf_{m}$ =$\,\{\,\,{Summfact_{m,1}\,,\dots,\,Summfact_{m,n}}\,\,\}\, $ \newline
\textbackslash \textbackslash Relevant Medical facts from the generated summary of query 'm'.
        
\vspace{-0.025cm}

\STATE $\#CorrectFacts_{m} = \lvert F_{m} \cap Sf_m \rvert$ 
\vspace{-0.025cm}
\IF{$\left(\text{Correct Medical Disorder phrase} \in \{ F_{m} \cap Sf_m \} \right)$}
    \STATE $\#CorrectFacts_{m} += 2$
\ELSIF{$ \left( \text{Partially correct disorder phrase} \in \{ F_{m} \cap Sf_m \} \right)$}
    \STATE $\#CorrectFacts_{m} += 1$
\ELSIF{$ \left( \text{Incorrect disorder phrase} \in \{ F_{m} \cap Sf_m \} \right)$}
    \STATE $\#CorrectFacts_{m} += -1$
\ELSE
    \STATE $\#CorrectFacts_{m} += 0$
\ENDIF

\begin{comment}
\State \textbf{if} $\, \left (\,{Correct Medical Disorder phrase\, \in \,\,\{\, F_{m} \,\cap Sf_m \} } \right )$: \newline
      \State \, \#\,CorrectFacts_{m} +=2 \newline
    \State \textbf{else if} $\,\left (\,{Partially correct disorder phrase \in \, \{\, F_{m} \, \cap\, Sf_m \, \}\, } \, \right)\,$: \newline
      \hspace{2cm}   \State \#CorrectFacts_{m} +=1 \newline
    \State \textbf{else if }$\,\left (\, {Incorrect disorder phrase \in \,\{\, F_{m} \cap Sf_m \,\}\, }\,\right)\,$: \newline
      \hspace{2cm}   \State \,\#\,CorrectFacts_{m} +=-1 \newline
    \State \textbf{else}: \newline
         \State  \,\#\,CorrectFacts_{m} +=0 \newline
%    \If{$i\geq 5$} 
%     \State $i \gets i-1$
% \Else
%     \If{$i\leq 3$}
%         \State $i \gets i+2$
%     \EndIf
% \EndIf 
\end{comment}  
  
\RETURN $\text{MMFCM} = \tanh(\#CorrectFacts_{m}/\lvert F_{m}\rvert)$
\end{algorithmic}
 \end{algorithm}

 \vspace{1cm}
\section{Qualitative Analysis}

We conducted a thorough qualitative analysis of the summaries generated by different models in both unimodal and multimodal setting. We also performed some case studies; one such instance is shown in figure \ref{fig:quali-ex1}. The analysis led to the following conclusion : (a) All models perform better in multimodal setting and is better at capturing the important visual information conveyed through the images and predicting the exact disorder phrase. (b) It was also observed that models like LLAMA-2 and Vicuna shows a tendency of hallucination (see figure \ref{fig:quali-ex1}) and sometimes generated facts which was completely out of context. The summaries generated by Mistral and Zephyr were more cohesive and showed lesser hallucinations. (c) FlanT5 showed the weakest results which made us conclude that decoder only LLMs perform better than LLMs with encoder-decoder architecture for our task.
% \vspace{-0.82cm}

\section{Risk Analysis}

We must acknowledge several limitations in our approach. Initially, we confined our task to 18 symptoms conducive to image sharing. Therefore, introducing an image outside this scope may lead to the model generating potentially erroneous information in the summary. While our multimodal model shows promise, it is prudent to engage a medical expert for the ultimate verification, particularly in high-stakes scenarios. Our AI model serves as a tool, not a substitute for medical professionals. 

\section{Conclusion and Future Work}

In this study, we delve into the impact of incorporating multi-modal cues, par-
ticularly visual information, on question summarization within the realm of
healthcare. We present the MMCQS dataset the first of its kind dataset, compris-
ing 3015 multimodal medical queries in Hindi-English codemixed language with
golden summaries in English that merge visual and textual data. This novel col-
lection fosters new assessment techniques in healthcare question summarization.
We also propose a model \textit{MedSumm}, that incorporates the power of both LLM
and Vision encoders to generate the final summary. In our future endeavors, we
aspire to develop a Vision-Language model capable of extracting the intensity and
duration details of symptoms and integrating them into the patient query’s final
summary generation. Furthermore, our expansion plans encompass incorporating
medical videos and speeches data and addressing scenarios involving other low
resources languages in Indian context.

\begin{comment}

This study explores the impact of incorporating visual information in healthcare question summarization, introducing the  MMCQS dataset. We present the \textit{MedSumm} model, merging LLM and Vision encoders. Future work includes developing a Vision-Language model for symptom intensity and duration extraction, integrating them into patient query summaries, and expanding into medical videos, speech data, and low-resource Indian languages.
\end{comment}

\section{Ethical Considerations}
\vspace{-0.1cm}
In healthcare summarization, we prioritize ethical considerations, including safety, privacy, and bias. We took extensive measures with the MMCQS dataset, collaborating with medical professionals, obtaining IRB approval, and adhering to legal and ethical guidelines during data handling, image integration, and summary annotation. The dataset is based on the HealthcareMagic Dataset, and medical experts supervised the task. Identity protection was ensured for user privacy.

\section{Acknowledgements}
Akash Ghosh and Sriparna Saha express their heartfelt gratitude to the  SERB (Science and Engineering Research Board ) POWER scheme of the Department of Science and Engineering, Govt. of India, for providing the funding for carrying out this research

\bibliography{MMCQS}
\bibliographystyle{abbrv}

\end{document}